\documentclass{article}

\usepackage{spconf,amsmath,graphicx}
\usepackage{booktabs}
\usepackage{caption}
\newcommand{\norm}[1]{\left\| #1 \right\|_{0}}
\usepackage{booktabs}
\usepackage{tabularx}
\usepackage{xcolor}
\usepackage{balance}
\usepackage{soul}
\usepackage{cite}
\usepackage{amssymb}
\usepackage{eso-pic}
\usepackage{subcaption}
\usepackage{tikz}
\usepackage{scrlayer-scrpage}
\usepackage{hyperref}
\usepackage{bm}
\usepackage{caption}
\hypersetup{
    colorlinks=true,
    linkcolor=blue,
    citecolor=green,
    filecolor=magenta,
    urlcolor=cyan}

\renewcommand{\paragraph}[1]{\noindent\textbf{#1}\quad}

\title{Cost Aware Untargeted Poisoning Attack against Graph Neural Networks\footnote{2023 IEEE. Personal use of this material is permitted. Permission from IEEE must be obtained for all other uses, in any current or future media, including reprinting/republishing this material for advertising or promotional purposes, creating new collective works, for resale or redistribution to servers or lists, or reuse of any copyrighted component of this work in other works.}}
\name{Yuwei Han, Yuni Lai, Yulin Zhu, Kai Zhou}
\address{Department of Computing, Hong Kong Polytechnic University, HKSAR\\ \{22041949r, yunie.lai\}@connect.polyu.hk, \{yulin.zhu, kai.zhou\}@polyu.edu.hk}

\begin{document}
\pagestyle{empty}

%
\maketitle
\begin{abstract}
Graph Neural Networks (GNNs) have become widely used in the field of graph mining. However, these networks are vulnerable to structural perturbations. While many research efforts have focused on analyzing vulnerability through poisoning attacks, we have identified an inefficiency in current attack losses. These losses steer the attack strategy towards modifying edges targeting misclassified nodes or resilient nodes, resulting in a waste of structural adversarial perturbation. To address this issue, we propose a novel attack loss framework called the Cost Aware Poisoning Attack (CA-attack) to improve the allocation of the attack budget by dynamically considering the classification margins of nodes. Specifically, it prioritizes nodes with smaller positive margins while postponing nodes with negative margins. Our experiments demonstrate that the proposed CA-attack significantly enhances existing attack strategies.

\end{abstract}
\begin{keywords}
Poisoning attack, graph neural networks, node classification.
\end{keywords}

\section{Introduction}
\label{sec:intro}
Graph Neural Networks (GNNs)~\cite{Jin,kipf2017semi,veličković2018graph} have emerged as an effective machine learning approach for structured data, generalizing neural networks to manage graph-structured information. They have shown potential in various applications such as social network analysis~\cite{Fan}, anomaly detection~\cite{Cha,pmlr-v162-tang22b,Duan}, and natural language processing~\cite{MAL}, excelling in tasks like node classification and link prediction. However, current research~\cite{pmlr-v80-dai18b,Wang} indicated that they are susceptible to \textit{poisoning attacks}, where the attack easily manipulates the graph structure (i.e., adding/deleting edges) before the training of the GNN models, and leads to incorrect predictions. Such vulnerabilities not only compromise the model's reliability but also present opportunities for malicious attacks. 

To analyze the vulnerabilities of GNNs, several studies have investigated potential poisoning attacks on GNNs, such as Minmax~\cite{inproceedings}, Metattack~\cite{zügner2018adversarial}, and certify robustness inspired attack framework~\cite{DBLP}. These attacks leverage gradient-based techniques to approximate the complex bi-level optimization problem associated with poisoning attacks. Nonetheless, there is still potential for enhancing the current methodologies regarding the allocation of the limited perturbation budget. 

We have noted Metattack with current loss functions such as negative log likelihood loss and CW loss~\cite{carlini2017towards} has budget waste problem that some budgets do not contribute to decrease the classification accuracy. For example, when employing the commonly used negative log likelihood loss as the attack objective, a node that has already been successfully misclassified tends to attract higher meta-gradients, resulting in a further allocation of the attack budget to that specific node. Wang et al.~\cite{DBLP} have proposed a novel attack loss framework called the Certify Robustness Inspired Framework (CR-framework) to enhance the attack performance. This method assigns larger meta-gradients to nodes with smaller certify robustness sizes ~\cite{DBLP}, as these nodes are more vulnerable to attacks. It accomplishes this by reweighting the loss of each node according to its certify robustness size. While this approach optimizes the budget allocation for existing attack losses and further decreases the accuracy of victim models, it grapples with computational inefficiencies during the assessment of certified robustness sizes.




To address the limitations above, we introduce a \textbf{Cost Aware Poisoning Attack Loss Framework (CA-attack)} to \textit{improve the allocation of attack budget} to maximize the impact of the attack. Specifically, we dynamically reweight the nodes according to their classification margins in the attack loss. This means that the weights of the nodes are adjusted during the optimization process of the attack objective. 
Through extensive testing on benchmark datasets, the CA-attack consistently surpasses previous attack methods in terms of effectiveness. Our research makes the following contributions:
\begin{itemize}

\item To address the inefficiencies of
budget allocation, we propose the CA-attack, a budget-efficient attack loss framework.
\item Through rigorous empirical assessments on three datasets, we demonstrate that CA-attack improves existing methods, highlighting its potential as a plug-and-play solution for various graph poisoning attacks.
\end{itemize}

\section{Preliminaries}
\label{sec:FORMULATION}
\paragraph{GNNs}We define an undirected graph as $G = \left(V, E \right)$, where $V=\{v_1,v_2,\cdots,v_N\}$ represents the node set, and $ e_{ij}\in E$ is edge connecting the nodes $v_i$ and $v_j$. Node attributes are represented by $X \in \mathbb{R}^{N\times d}$, where $d$ is the dimension of the attribute. Additionally, the graph structure can be represented by an adjacency matrix $A\in \{0,1\}^{N\times N}$, with $A_{ij} = 1$ if an edge exists between two nodes, otherwise $0$. Given a subset of labeled nodes $V_L\subset V$ where each node $v\in V_L$ has a label $y_v\in \mathcal{C}=\{c_1,c_2,\cdots,c_k\}$, GNNs aim to learn a function $f_{\theta}$ to predict the remaining unlabeled nodes $V_U=V\setminus V_L$ into classes of $\mathcal{C}$. We use $f_{\theta}(G)_v$ to denote the prediction of the model $f_{\theta}$ for node $v$. Its parameters  $ \theta $  are optimized by minimizing a loss $ \mathcal{L} _{train}$ over the labeled nodes, typically using losses like negative log-likelihood loss or CW loss.\\
\paragraph{Attacks against GNNs}Based on the classification task, the general form of node-level graph adversarial attacks can be defined as a bi-level optimization process where the attacker optimizes the allocation of the attack budget to maliciously modify the graph, while the model is optimized using this poisoned graph:
\begin{eqnarray}
\begin{aligned}
	\min    \mathcal{L} _{atk} \left( f_{\theta ^*}\left( \widehat{G} \right) \right) =\sum_{v\in V}{\ell _{atk}\left(f_{\theta ^*}\left( \widehat{G} \right) _v,y_v \right)}\\
       s.t., \theta ^*=\underset{\theta}{arg\min}\mathcal{L} _{train}\left( f_{\theta}\left( \widehat{G} \right) \right),
\end{aligned} 
\label{f1}
\end{eqnarray}
where $ \ell _{atk} $ is the attack loss and usually chosen as $ \ell _{atk} = - \mathcal{L} _{train}$. $\widehat{G} $ is the graph modified from original $ G $ by the adversarial attack. $ y_v$ denotes the labels of node $ v $.

Given a budget $\Delta$, the attacker aims to ensure that perturbations are unnoticeable by maintaining the $ l_0 $ norm difference between the original and perturbed graph:
$\norm{A-A^{\prime}} \leq \Delta$.
To avoid detection, the attacker also refrains from making significant changes to the graph's degree distribution or introducing isolated nodes, as suggested in \cite{zugner2018adversarial}.

A conventional poisoning attack process can be divided into three steps. The first step is to retrain a surrogate model $f_{\theta ^*}\left( G \right)$ using a linearized graph convolutional network (GCN) which is expressed as:
\begin{eqnarray}
	f_{\theta} \left(G \right) = soft\max \left( \hat{A}^2XW \right),
	\label{f2}
\end{eqnarray}
 where $\hat{A}=D^{-1/2}\left( A+I \right) D^{-1/2}$ is the normalized adjacent matrix, $X$ are the node features, $D$ is the diagonal matrix of the node degrees, and $\theta=\left\{W \right\}$ are the set of learnable parameters. In the second step, the attacker uses pseudo-labels $y_v$ generated by the surrogate model $f_{\theta ^*}\left( G \right)$ to construct the attack loss of the node $v$ as $\ell\left(f_{\theta ^*}\left( \widehat{G} \right) _v,y_v \right)$ under the gray-box attack scenario. In the third step, the attack loss of each node is backpropagated to produce a partial gradient matrix:
 \begin{eqnarray}
	g_v = \nabla _A\ell\left(f_{\theta ^*}\left( \widehat{G} \right) _v,y_v \right),
	\label{f3}
\end{eqnarray}
 The overall gradient information passed to the attack strategy is the average of all the partial gradient matrices. Thereafter, the attacker selects the edges to be perturbed (adding/deleting) based on the saliency of their gradients.
\section{Cost-Aware Attack}
\label{sec:attack}
In this section, we demonstrate our proposed CA-attack attack loss framework that aims to improve the approximation of this challenging bi-level optimization problem~\eqref{f1} by redesigning the $\mathcal{L} _{atk}$. Specifically, we incorporate the classification margin of nodes as dynamic weights for the attack loss. 

\subsection{Limitations of Previous Attacks}
\begin{figure}[t]
    \centering
    
    \begin{subfigure}[b]{0.3\linewidth}
        \centering
         \hspace{1cm}$\mathcal{L}_{CE}$\\[1ex] 
        \begin{tikzpicture}
            \node[anchor=south west,inner sep=0] (image) at (0,0) {\includegraphics[width=\linewidth]{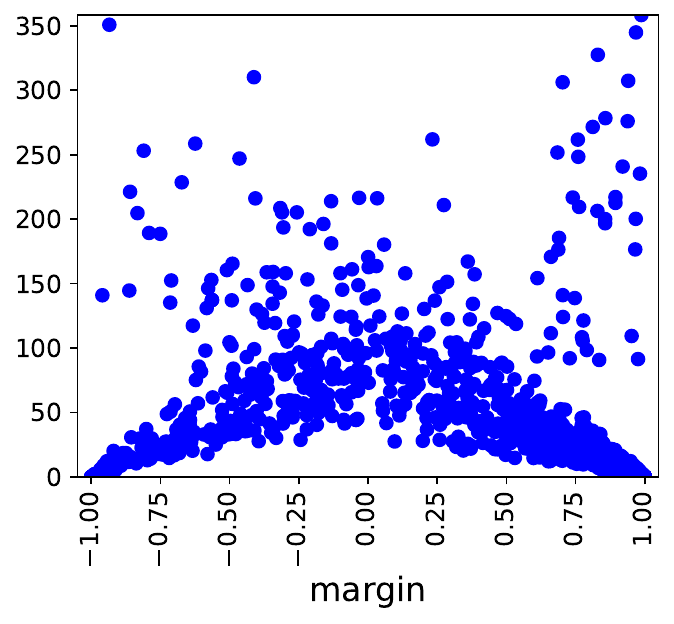}};
             \node[rotate=90, anchor=center, yshift=0.125cm,font=\tiny\bfseries] at (0,0.6*\linewidth){$\left\|\bm{g_v}\vphantom{\rule{1pt}{1pt}} \right\| _2$};
             
        \end{tikzpicture}

    \end{subfigure}
    \hfill 
    \begin{subfigure}[b]{0.3\linewidth}
        \centering
        $\mathcal{L}_{CR}$\\[1ex]
        \includegraphics[width=\linewidth]{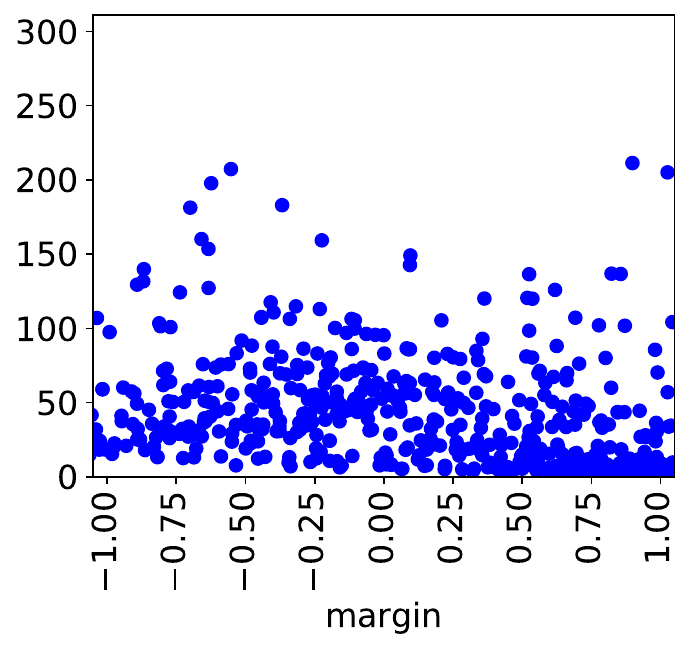}
        
    \end{subfigure}
   \hspace{-0.04cm}
    \begin{subfigure}[b]{0.3\linewidth}
        \centering
        $\mathcal{L}_{CA}$\\[1ex]
        \includegraphics[width=\linewidth]{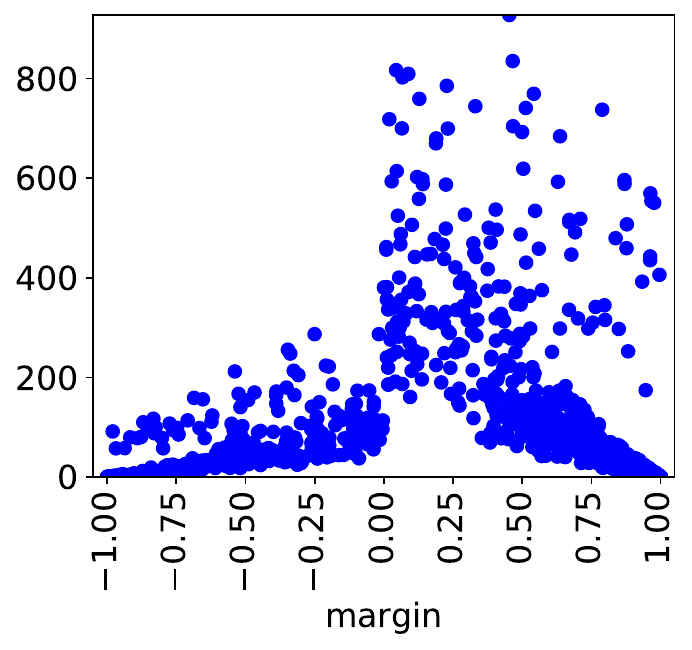}
        
    \end{subfigure}

    \caption{Scatterplot of the nodes’ margins with $l_2$ norm of their partial gradient matrices of cora dataset. The terms $\mathcal{L}_{CE}$, $\mathcal{L}_{CR}$~\cite{DBLP}, and $\mathcal{L}_{CA}$ respectively represent the negative log likelihood loss, the CR framework, and the CA framework.}
    \label{fig:res1}
\end{figure}

Since poisoning attack aims to minimize the attack loss (i.e., maximize training loss) by modifying an edge with the largest gradient at each iteration, the node connected to this edge will be attacked with higher priority according to Equation~\eqref{f3}. Figure \ref{fig:res1} illustrates the priority of a node in a clean graph to be attacked based on the $l_2$ norm of partial gradient matric for each node in Metattack using different attack loss functions. We can observe that the negative log-likelihood loss and the CR framework result in significant gradients for nodes with negative margins. Additionally, the negative log-likelihood loss produces significant gradients for nodes with large margins. In contrast, our CA framework generates significant gradients on nodes with small positive margins. In Section~\ref{sec:ex}, we will demonstrate that the CA attack loss outperforms previous attack losses.

\subsection{Cost Aware Loss}
The classification margin of a node $v$ is commonly defined as:
\begin{eqnarray}
	\varphi(v) = z_{c^*}-\max _{c\ne c^*}z_c,
	\label{f4}
\end{eqnarray}
where $z$ represents the vector of logits produced by the model towards node $v$, and $c^*$ refers to the true label of the node $v$. 
If a node has a negative margin, it indicates misclassification. Intuitively, \textit{a larger margin suggests that a node is more resilient to attacks, whereas a smaller but positive margin suggests a higher potential for a successful attack.}

With this intuition in mind, we propose the introduction of a \textbf{cost-aware loss (CA-loss)} that takes into account the margins of nodes. By doing so, our aim is to assign higher priority to nodes with smaller but positive margins.



\begin{figure}[t]
    \centering
    \includegraphics[width=0.5\linewidth]{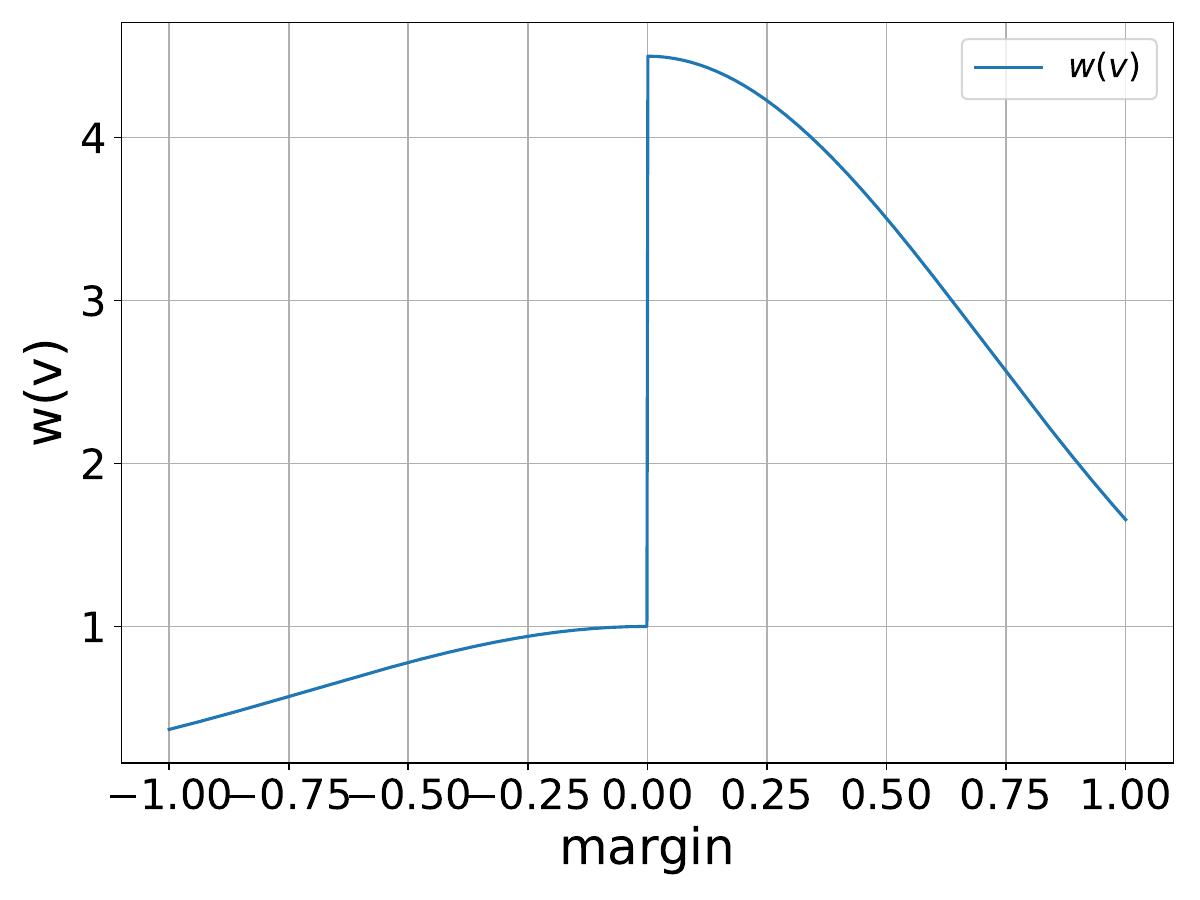}
    \caption{Weights of nodes in Cora dataset with different margins.}
    \label{fig:res2}
\end{figure}

While a simplistic strategy is to rank the nodes based on margins and perturb them sequentially, this approach is computationally demanding and potentially suboptimal due to the complex interplay of nodes and edges in predictions. Instead, we refine the attack loss $\ell_{atk}$ by incorporating node margins to weight the nodes dynamically. The resulting CA-loss is defined as follows:
\begin{eqnarray}
\begin{aligned}
	\mathcal{L}_{CA}\left( f_{\theta ^*},\widehat{G}\right) =\sum_{v \in V_U}^{}{w\left(v \right) \cdot \ell\left(f_{\theta ^*}\left( \widehat{G} \right) _v,y_v \right)},
\end{aligned} 
\label{f5}
\end{eqnarray}
where $w\left(v \right)$ is the weight of the node $v$. When all nodes are assigned equal weight, our CA-loss reduces to the conventional loss. To introduce the weight $w(v)$, which captures the inverse relationship between the node's margin $\varphi(v)$ and the weight, we utilize the exponential function. The weight $w(v)$ is defined as follows:
\begin{eqnarray}
\begin{aligned}
	w(v) = \alpha\times e^{-\beta \times \varphi (v)^2},
\end{aligned} 
\label{f6}
\end{eqnarray}
where $\alpha$ and $\beta$ are tunable hyper-parameters. The weight $w(v)$ exponentially decreases as the node's margin $\varphi(v)$ increases. Figure \ref{fig:res2} is a visualization of $w(v)$. This property ensures that the majority of perturbed edges are utilized to disrupt nodes with smaller margins during the attack process.


\section{Experiments}
\label{sec:ex}
In this section, We present the experimental results by answering the following questions:
\begin{itemize}
    \item $RQ1$:Can our CA attack loss framework enhance the efficacy of current poisoning attacks? 
    \item $RQ2$:Can our CA attack loss framework serve as a universal framework for conventional poisoning attack losses?
    \item $RQ3$:Is our framework more computationally efficient compared to baselines?
\end{itemize}
\begin{table}[t]
    \centering
    \caption{Dataset statistics.Following~\cite{Jin}, we only consider the largest connected component (LCC).}
    \begin{tabular}{@{}cccccc@{}}
    \toprule
   
         \textbf{Dataset}& \textbf{Nodes} & \textbf{Edges}& \textbf{Classes}& \textbf{Features} \\
        
		\midrule
		Cora& 2,485& 5,069 & 7 & 1,433 \\
	
		Citeseer& 2,110& 3,668 & 6 & 3,703  \\
            Polblogs & 1,222& 16,714 & 2& $/$  \\
  \bottomrule
    \end{tabular}
   
    \label{Table1}
\end{table}
\begin{table}[t]
    \centering
    \caption{Hyperparameters of CA.}
    \begin{tabular}{@{}lcccc@{}}
    \toprule
        \textbf{Dataset} & \textbf{$\alpha1$}& \textbf{$\alpha2$} & \textbf{$\beta1$}& \textbf{$\beta2$} \\
	\midrule
		Cora & 4.5 & 1.0 & 1.0 & 1.0 \\
		Citeseer & 4.5 & 1.0 & 1.0 & 1.0 \\
        Polblogs & 1.0 &1.0 & 0.5 & 0.1 \\
    \bottomrule
    \end{tabular}
    \label{Table2}
\end{table}
\begin{table*}[t]
\centering
\caption{Experimental results of attacks for GCNs. The best results from each experiment are bold.}

\begin{tabular}{@{}llllllllll@{}}
\hline

Dataset&\multicolumn{2}{c}{Cora} &
\multicolumn{2}{c}{Citeseer} & 
\multicolumn{2}{c}{Polblogs} 
 \\
\hline
Pert Rate(\%)   & \multicolumn{1}{c}{5\%} &      \multicolumn{1}{c}{10\%} & \multicolumn{1}{c}{5\%}&           \multicolumn{1}{c}{10\%}& \multicolumn{1}{c}{5\%}  &     \multicolumn{1}{c}{10\%}\\

\hline
Clean       & \multicolumn{2}{c}{84.61$\pm$0.28}     &   \multicolumn{2}{c}{73.95$\pm$0.28}   &   \multicolumn{2}{c}{95.34$\pm$0.19}        \\
\hline
DICE        &  82.79$\pm$0.17      &   81.92$\pm$0.22     &  72.32$\pm$0.24    &  71.59$\pm$0.28   &   86.54$\pm$0.42       &   80.71$\pm$0.29        \\

CE-MetaSelf   &  76.56$\pm$0.27   &   66.96$\pm$0.55   &  71.49$\pm$0.39    &  66.20$\pm$0.30   &   77.22$\pm$0.34       &   69.73$\pm$0.44          \\

CR-CE-MetaSelf & 75.79$\pm$0.34    &  68.50$\pm$0.29    &   65.28$\pm$0.43   & 56.93$\pm$0.33    &  78.68$\pm$0.32        &    70.43$\pm$0.85    \\

\textbf{CA-CE-MetaSelf} &  \textbf{66.98$\pm$0.56}  &   \textbf{56.30$\pm$0.42}   & \textbf{62.58$\pm$0.28}    &  \textbf{53.22$\pm$0.30}   &   \textbf{76.58 $\pm$0.46}     &   \textbf{69.43$\pm$0.22}          \\
\hline\hline
CW-MetaSelf      &  72.75$\pm$0.50     &  61.85$\pm$1.18    &  64.37$\pm$1.07   & \textbf{52.17$\pm$0.25}   &  83.71$\pm$0.66        &  79.45$\pm$0.47        \\

CR-CW-MetaSelf    & 71.40$\pm$0.61 & 60.84$\pm$0.45   &  64.76$\pm$0.88  & 54.98$\pm$1.29   & \textbf{83.09$\pm$0.92}     &       79.87$\pm$1.02     \\

\textbf{CA-CW-MetaSelf} &  \textbf{69.69$\pm$0.64}   &   \textbf{58.50$\pm$0.73}   &  \textbf{60.79$\pm$0.51}    & 53.28$\pm$0.62    &   83.24$\pm$0.45       &    \textbf{76.99$\pm$0.46}        \\

\hline
\end{tabular}
 \label{Table3}
\end{table*}


\begin{table}[t]
\centering
\caption{Comparison of computational efficiency.}
\begin{tabular}{@{}lllll@{}}
\hline
Methods     & Cora & Citeseer & Polblogs \\
\hline
DICE        &  0.04s    &  0.01s        &  0.06s        \\
CE-MetaSelf    &   107.46s   &   97.90s       &      114.07s    \\
CW-MetaSelf    &   158.39s   &   129.12s       &      236.52s    \\
CR-CE-MetaSelf &  84533.67s    &   53089.36s       &   42780.71s       \\
CR-CW-MetaSelf   & 87566.54s     & 56345.78s         & 53390.56s         \\
\textbf{CA-CW-MetaSelf}   &  153.58s    &   122.74s       &  200.68s        \\
\textbf{CA-CE-MetaSelf} &  104.39s    &   107.88s       &   112.37s       \\
           
\hline
\end{tabular}
 \label{Table5}
\end{table}

\label{sec:pagestyle}
\subsection{Set Up}
\label{ssec:subhead}
\paragraph{Datasets and GNN models.}We evaluate our attacks on three benchmark graph datasets, i.e., Cora~\cite{McCall}, Citeseer~\cite{Sen} and Polblogs~\cite{article}. Table \ref{Table1} shows the basic statistics of these datasets. The datasets are partitioned into 10\% labeled nodes and 90\% unlabeled nodes. The true labels of the unlabeled nodes are hidden from both the attacker and the surrogate, serving only as a benchmark for assessing the effectiveness of the adversarial attacks. We employ the Graph Convolutional Network (GCN)~\cite{kipf2017semi}  as our target GNN model.\\
\paragraph{Baseline Attacks.}We choose DICE~\cite{Wan}\footnote{\url{https://github.com/DSE-MSU/DeepRobust}}, MetaSelf~\cite{zügner2018adversarial}\footnote{\url{https://github.com/DSE-MSU/DeepRobust}} and Certify robustness(CR) inspired attack framework~\cite{DBLP} as the base attack methods. Details of baselines are discussed in Section~\ref{sec:intro}. For clarity in presentation, we denote the variant of MetaSelf utilizing the negative log likelihood loss as CE-MetaSelf. When employing the CW loss~\cite{carlini2017towards}, it is referred to as CW-MetaSelf. When integrating the CR-inspired attack framework with MetaSelf using the negative log likelihood loss, we term it CR-CE-MetaSelf. Similarly, its adaptation with the CW loss is named CR-CW-MetaSelf.\\
\paragraph{Implementation Details}All attacks are implemented in PyTorch and run on a Linux server with $16$ core $1.0$ GHz CPU, $24$ GB RAM, and $10$ Nvidia-RTX $3090$Ti GPUs. 
Each experiment is repeated for ten times under different initial random seeds, and the uncertainties are presented by $95\%$ confidence intervals around the mean in our tables.
We separately set the perturbation budget $\Delta$ as $5\%$ and $10\%$ of the total number of edges in a graph. The specific parameters associated with our cost aware loss were selected by grid search to achieve optimal attack performance and details are in Table \ref{Table2}. For $\varphi (v) >0$, we use $\alpha1$ and $\beta1$. Conversely, for $\varphi (v) <0$, we employ $\alpha2$ and $\beta2$. For the baseline CR inspired attack framework, we adopted the parameter settings as outlined in their original paper and computed the certify robustness size at every $20th$ iteration during the generation of the poisoned graph.
\subsection{Results and Analysis}
\paragraph{Performance Comparison($RQ1$)}Table \ref{Table3} presents a comparison of our attack performance against other established methods. We can observe that our CA attack loss framework combined with negative log likelihood loss can  enhance the CE-MetaSelf performance in all datasets. For instance, when attacking GCN with a 5\% perturbation ratio, our CA-CE-MetaSelf demonstrates a gain of 9.58\% and 9.74\% over CE-MetaSelf on Cora and Citeseer, respectively. Moreover, CA-CE-Metaself has a relative improvement of 12.20\% and 3.71\% compared to CR-CE-MetaSelf, with a perturbation ratio of 10\% on Cora and Citeseer, respectively. These findings indicate that our CA attack loss framework enhances poisoning attack performance more effectively within the same budget constraints.\\
\paragraph{Universality Analysis($RQ2$)}Table \ref{Table3} highlights the effectiveness of our weight assignment strategy when applied to the CW loss, underscoring the versatility of our approach. For example, when considering a 5\% perturbation ratio, our proposed CA-CW-MetaSelf approach exhibits a relative gain of 3.58\% and 3.06\% over the CW-MetaSelf method for the Citeseer and Cora datasets, respectively. Additionaly, CW-CE-MetaSelf has a gain of 1.71\% and 3.97\% over the CR-CW-MetaSelf for the Cora and Citeseer datasets, respectively. Since we use the same parameters with negative log likelihood loss for CW loss to reduce the time for parameter adjustments, it may not be optimal for the CW loss which could result in suboptimal results compared to the baselines. The results of our attack can be improved with more detailed parameter tuning for the CW loss within the CA attack loss framework. \\
\paragraph{Computational Efficiency($RQ3$)}We compared the computational efficiency of our method with the baselines, shown in Table \ref{Table5}. DICE has the highest computational efficiency since it depends on randomness without gradient derivation. The computational efficiency of CR-CE-MetaSelf and CR-CW-MetaSelf are the lowest because they need much time to compute certify robustness size for every node. CA-CE-MetaSelf and CA-CW-MetaSelf have similar computational efficiency with CE-MetaSelf and CW-MetaSelf. Therefore, the computational efficiency of CA attack loss framework is satisfactory while the performance is competitive.\\

\label{ssec:subhead}
\section{Conclusion}
\label{sec:typestyle}

We investigate graph poisoning attacks on graph neural networks and present an innovative Cost Aware Poisoning Attack Loss Framework(CA-attack) utilizing the concept of classification margin. Our research begins by examining the budget inefficiencies present in previous
attack methods. Subsequently, we develop a cost-aware loss framework that assigns dynamic weights to victim nodes based on their margins. Through evaluations conducted on various datasets, we demonstrate that our attack loss can considerably decrease the budget waste and improve the performance of existing attacks.\\

\textit{Note:The copyright of this paper is reserved by 2023 IEEE.}




\newpage

\bibliographystyle{IEEEbib}
\bibliography{refs}

\end{document}